\documentclass[conference]{IEEEtran}
\IEEEoverridecommandlockouts
\usepackage[utf8]{inputenc} % allow utf-8 input
\usepackage[T1]{fontenc}    % use 8-bit T1 fonts
\usepackage{cite}
\usepackage{amsmath,amssymb,amsfonts}
\usepackage{algorithmic}
\usepackage{graphicx}
\usepackage{textcomp}
\usepackage{xcolor}
\usepackage{booktabs}
\usepackage{multicol}
\usepackage{hyperref}
\usepackage{subfig}
\usepackage{balance}
\usepackage{todonotes}
\usepackage{authblk}

\def\BibTeX{{\rm B\kern-.05em{\sc i\kern-.025em b}\kern-.08em
    T\kern-.1667em\lower.7ex\hbox{E}\kern-.125emX}}
\begin{document}
\bstctlcite{IEEEexample:BSTcontrol}

\newcommand\copyrighttext{%
  \footnotesize \textcopyright 2022 IEEE. Personal use of this material is permitted.
  Permission from IEEE must be obtained for all other uses, in any current or future
  media, including reprinting/republishing this material for advertising or promotional purposes, creating new collective works, for resale or redistribution to servers or lists, or reuse of any copyrighted component of this work in other works. DOI: \href{http://doi.org/10.1109/IPDPSW55747.2022.00126}{10.1109/IPDPSW55747.2022.00126}}
\newcommand\copyrightnotice{%
\begin{tikzpicture}[remember picture,overlay]
\node[anchor=south,yshift=7pt] at (current page.south) {\fbox{\parbox{\dimexpr\textwidth-\fboxsep-\fboxrule\relax}{\copyrighttext}}};
\end{tikzpicture}%
}

%TODO: working title
\title{A Green(er) World for A.I. \\
%Energy to Performance; 

\thanks{This material is based upon work supported by the Assistant Secretary of Defense for Research and Engineering under Air Force Contract No. FA8702-15-D-0001, and United States Air Force Research Laboratory Cooperative Agreement Number FA8750-19-2-1000. Any opinions, findings, conclusions or recommendations expressed in this material are those of the author(s) and do not necessarily reflect the views of the Assistant Secretary of Defense for Research and Engineering, or the United States Air Force. The U.S. Government is authorized to reproduce and distribute reprints for Government purposes notwithstanding any copyright notation herein.}
}

\author{
        Dan Zhao\IEEEauthorrefmark{1},
        Nathan C. Frey\IEEEauthorrefmark{1},
        Joseph McDonald\IEEEauthorrefmark{1},
        Matthew Hubbell\IEEEauthorrefmark{1}, \\
        David Bestor\IEEEauthorrefmark{1},
        Michael Jones\IEEEauthorrefmark{1},
        Andrew Prout\IEEEauthorrefmark{1},
        Vijay Gadepally\IEEEauthorrefmark{1},
        Siddharth Samsi\IEEEauthorrefmark{1}\textsuperscript{\textsection} \\
        \IEEEauthorrefmark{1} MIT Lincoln Laboratory}

%\affil[*]{dan.zhao@ll.mit.edu}

\maketitle

\begingroup\renewcommand\thefootnote{\textsection}
\footnotetext{Corresponding author. Email : \url{sid@ll.mit.edu}}
\copyrightnotice
\endgroup

\begin{abstract}
As research and practice in artificial intelligence (A.I.) grow in leaps and bounds, the resources necessary to sustain and support their operations also grow at an increasing pace. While innovations and applications from A.I. have brought significant advances, from applications to vision and natural language to improvements to fields like medical imaging and materials engineering, their costs should not be neglected. As we embrace a world with ever-increasing amounts of data as well as research \& development of A.I. applications, we are sure to face an ever-mounting energy footprint to sustain these computational budgets, data storage needs, and more. 
But, is this sustainable and, more importantly, what kind of setting is best positioned to nurture such sustainable A.I. in both research and practice? In this paper, we outline our outlook for Green A.I.---a more sustainable, energy-efficient and energy-aware ecosystem for developing A.I. across the research, computing, and practitioner communities alike---and the steps required to arrive there. We present a bird's eye view of various areas for potential changes and improvements from the ground floor of AI's operational and hardware optimizations for datacenter/HPCs to the current incentive structures in the world of A.I. research and practice, and more. We hope these points will spur further discussion, and action, on some of these issues and their potential solutions.

% This document is a model and instructions for \LaTeX.
% This and the IEEEtran.cls file define the components of your paper [title, text, heads, etc.]. *CRITICAL: Do Not Use Symbols, Special Characters, Footnotes, 
% or Math in Paper Title or Abstract.
\end{abstract}

\begin{IEEEkeywords}
Green AI, sustainable AI, energy efficiency
\end{IEEEkeywords}

\section{Introduction}

Issues of environmental sustainability and energy efficiency have come to center stage as global warming, climate concerns, and their consequences have permeated many aspects of our economy and society. In finance, sustainable investing has come to the fore where, in addition to traditional metrics of assessing risk, themes of environmental, social, and governance (ESG) have become important in evaluating financial and purpose-driven outcomes. Throughout the private sector, many companies have begun to re-examine and prioritize green power usage and resource development \cite{epa100} while governments have begun to invest heavily in clean energy and climate resilient infrastructure \cite{whitehouse}---the list goes on.

While traditional sources of carbon emissions from agriculture and transportation continue to contribute the lion's share of greenhouse gas emissions in the U.S., electricity usage from the operation of supercomputing and data centers are climbing with historical signs of compute costs and demand accelerating further in the years ahead \cite{computeaccel}. Estimates place datacenters' electricity consumption at 1\% of global electricity demand \cite{iea} with projections of electricity usage reaching 8\%-21\% of global demand by 2030 \cite{naturedatacenter}, though extrapolation of demand trends can be unreliable due to not accounting for new improvements in energy efficiency \cite{recalibratecosts}. However, even beyond the energy footprint from electricity consumption, these datacenters can take up significant amounts of water, either directly for cooling or indirectly for electricity generation, bearing a larger than expected environmental footprint---in the U.S., it is estimated that 20\% of datacenter servers' direct water footprint is sourced from moderately to highly stressed watersheds and 50\% of servers are at least partially supplied by power plants in water stressed areas \cite{footprintdata}. In addition to the energy footprint datacenter/HPC operations, embodied carbon costs \cite{metasusAI} such as those associated with manufacturing hardware for A.I. development and applications also matter, especially as hardware continues to advance. As such, the environmental footprint of A.I. may go beyond the costs represented by carbon emissions of datacenters/HPCs alone.

\begin{figure}[!h]
\centerline{\includegraphics[width=0.4\textwidth]{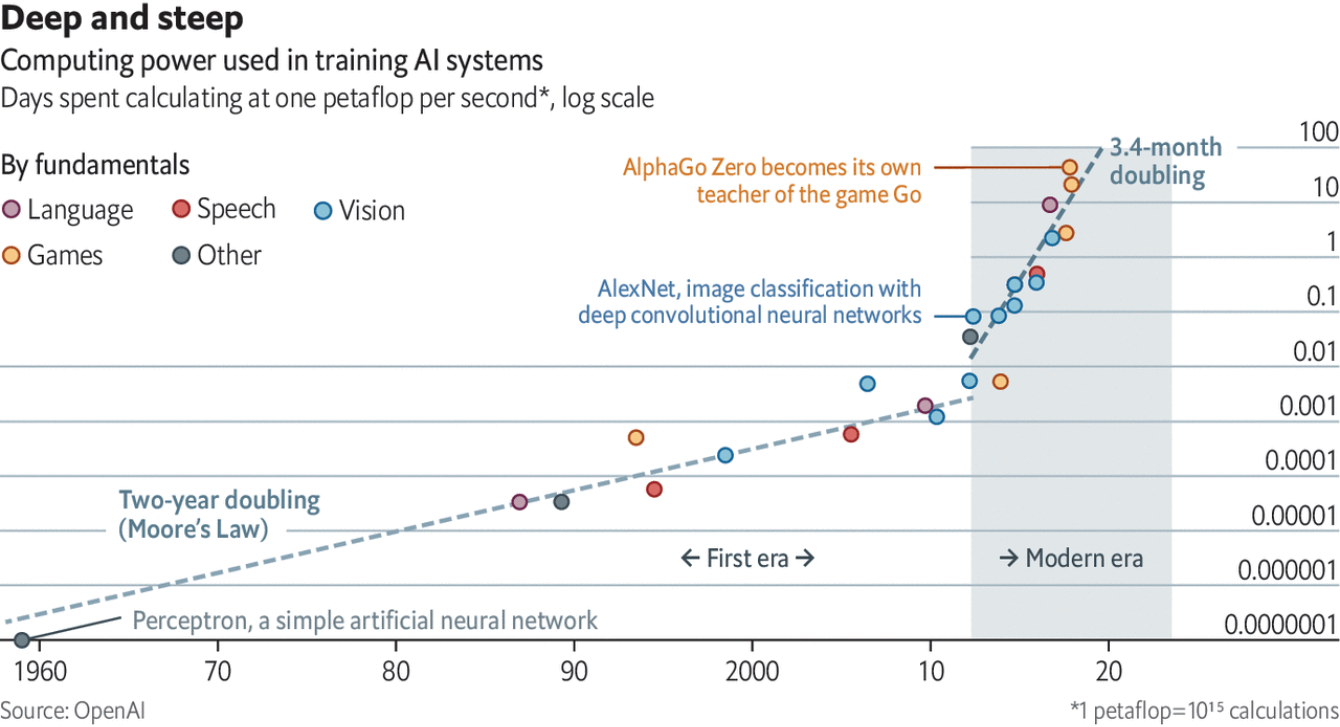}}
\caption{\small{\textbf{Modern AI's Computational Demands.} Note the steep increase in just the past decade relative to the past 50 years. \textit{Source: OpenAI \& The Economist.}}}
\label{fig:ai_computation}
\end{figure}

As industry adoption and incorporation of algorithms into products and services become more commonplace, we have seen significant growth in both the amounts of training data and the size of the model itself \cite{metasusAI} as the main means to realize performance gains. Simultaneously, fundamental A.I. research has continued to accomplish increasingly complex tasks with increasingly complex models and large datasets. These factors have, in turn, inevitably pushed similar growth trends in infrastructure investments required to keep pace with the increased amounts of training, inference, storage, and more (see Fig. \ref{fig:ai_computation}). Coupled with the anticipated increase in internet traffic, consumer devices, and demand for the very products and services some of these algorithms support \cite{naturedatacenter}, these worrying trends in energy demand and its associated energy footprint are likely to only accelerate. Even so, in the arms-race for A.I. superiority and operationalization, companies and institutions involved in A.I. research and its applications have continually expanded their datacenters and operations. Google's datacenter facilities span several countries while Meta has recently announced the construction of a new A.I. Research SuperCluster (RSC) claimed to be among the fastest and largest supercomputing centers upon completion\cite{metaAIcenter}. In this race to construct an ever-increasing number (and size) of datacenters, supercomputing clusters, and supporting facilities, there are few signs that this race will slow down. Instead, companies are accepting this as an inevitability and are looking for ways to help offset their ever-increasing energy footprint, such as building their own additional energy production facilities to fuel their operations \cite{googlesolar} \cite{microsoftwater}. 

Energy-efficient data infrastructure and green computing are hardly new concepts and have seen continued work and advances. From the development of efficient chips like Google's TPUs \cite{googletpu} and other computing efficiency gains to the application of A.I. algorithms themselves to automate datacenter operations, there is a long list of existing practices and current works-in-progress to address the energy-hungry and data-intensive appetite necessary to sustain these algorithms. Though these advances in efficiency have kept pace with the increased computation/energy needs and offset demand thus far, there may be signs that this is unlikely to last \cite{pue}. There also exists some debate on the true extent to which issues on A.I.'s sustainability and energy footprint are accurately described, largely driven by notable successes in realizing energy and computational efficiency in model training, datacenter/HPC operation, and hardware. However, changes in climate resulting in rising temperatures and more extreme weather patterns are likely to stress cooling and already strained resources in many areas. While larger, well-equipped technology companies have the resources and incentives to act, develop, and adopt efficiently, there are still clear, unaddressed concerns if all A.I. workflows move to the same hardware and software stack despite the efficiency benefits from centralization. As we run up against the limits of remaining efficiency gains, other ideas and implementations are needed, either as an anticipatory or preventative measure, in order to proactively develop strategies that bring the discourse to these problems and their potential solutions.

% Policy and market design solutions have also appeared in the form of mechanisms such as carbon offsets but their usage (and abuse) and their implementation have faced scrutiny over questions of monitoring and actual effectiveness in practice. 

In the following sections, we discuss the prospects of encouraging energy efficiency across various levels of the research \& development spectra of A.I. and its applications: (1) \textit{the infrastructure and resource utilization level}, (2) \textit{the individual user and behavioral level}, and (3) \textit{the group and community of A.I. researchers and practitioners at large}. These three aspects cover issues from a micro-to-macro perspective but also emphasize a key point---no single change on any one level is likely to be as effective without corresponding changes on the other levels since these three aspects are part of a single whole. A concerted, unified effort is required in order to transition effectively to a greener ecosystem for A.I. research and practice. To make our analyses more concrete in our discussions, we leverage data from the MIT SuperCloud \cite{reuther2018interactive}, an operational peta-scale HPC system that is actively used for research, experimentation, and collaborations by the MIT research community in several disciplines across machine learning, deep learning, and more.

% From the resource and infrastructure level, we examine and outline potential improvements relevant to aspects like optimization and management of HPCs/datacenters, their energy utilization and energy purchase patterns, system hardware optimizations at scale and user-centric/workload-specific configurations, data storage and growth, algorithmic/model choices, system hardware optimizations at scale, the measurement/tracking/evaluation of resource requirements and utilization across the full-stack development cycle (i.e. from experimentation/prototyping to training and inference), and more as they relate to the infrastructure and resource utilization required to sustainably power continued efforts in A.I. research and development. 

% On the individual user and behavioral level, we discuss how incentive structures and aspects of human behavior can interact to produce unintended effects, or result in very different effects on the aggregate level, and re-align to produce the intended effect. Lastly, on the community and group level, we examine the current state of the A.I. research ecosystem and its potential implications for aggregate energy utilization and computation demand, the incentives it induces, and the influence it bears on overall research priorities and research representation. We conclude with a brief discussion of fruitful directions to help push forward sustainable A.I. in research and in practice.

% \section{Outline}

\section{Energy \& Behavioral Considerations}
In this section, we discuss potential improvements towards a more energy-aware compute and cluster optimization framework. While we discuss traditional aspects of datacenter/HPC management in reducing energy expenditure (e.g. hardware, system-level), we also focus on non-traditional possibilities. We touch upon issues such as the economic considerations of energy consumption like the opportunity costs of energy purchases, the role/effect of user behavior in designing mechanisms to encourage energy-efficient behavior, changes in existing behaviors (e.g. from either the user side or datacenter/HPC management side), and combining---but balancing---existing energy saving mechanisms on the hardware/systems side with ones accounting for user behavior and incentives.

When it comes to energy efficiency, a simplified optimization framework is useful in understanding the objectives, the available choices/mechanisms are at our disposal to affect change, their dependence on one another, trade-offs, and more. This way, we can simplify the overall optimization problem that operational datacenters/HPCs face:
\begin{equation}
\label{energy_min}
\min_{q_{s}, p, c} E( q_d, q_s, p, c, \varepsilon) \hspace{2mm} \text{s.t.}  \hspace{2mm} A(q_d, q_s, p, c, \varepsilon) \geq \alpha
\end{equation}
where total energy expenditure $E(\cdot)$ and activity level $A(\cdot)$ of the datacenter/HPC can be affected by various factors: examples include the ``quantity'' of compute resources demanded or currently utilized by users  ($q_d$) as well as their usage behaviors including but not limited to efficient/inefficient practices, the ``quantity'' of compute resources supplied or available to users ($q_s$) and associated resource settings, the job scheduling system or resource allocation rule in place ($p$), control mechanisms ($c$) such as hardware settings (e.g. power caps, clock rate settings) or other physical interventions (e.g. rack placements, cooling setups) and ``softer'' mechanisms (e.g. algorithmic, instrumentation) that may be in place, and $\varepsilon$ which accounts for other factors such as temperature (e.g., ambient, distributions across racks, local climate) and others (e.g. a datacenter's fuel mix and energy purchasing patterns, maintenance schedules, electricity prices and energy mix). 

In other words, the goal is to minimize the energy expenditure $E(\cdot)$ of the datacenter subject to a constraint: the activity or performance level $A(\cdot)$ of the supercluster must be above some minimum, acceptable threshold $\alpha$. This constraint expresses a fundamental trade-off at the heart of energy-efficiency: reductions in energy consumption or expenditure need to be weighed against trade-offs in performance (i.e. jobs still need to be done at a reasonable pace). If the performance level constraint $\alpha$ is not satisfied, attempts to reduce energy expenditure may produce perverse, unintended effects; for instance, if a change to reduce energy consumption results in noticeable performance degradation, then users may run more jobs for longer, producing the opposite effect. Although one possibility is that higher throughput jobs can reduce total energy consumption by driving up power consumption but finishing in shorter periods as a result, we assume here that $\alpha$ corresponds to a bare minimum performance level---beneath which even these high throughput jobs contribute little to the overall energy footprint compared to the other kinds of workloads/operations present.

Traditionally, resource management in datacenters/HPCs tends to take an approach closely aligned with the problem as outlined (Eq. \ref{energy_min}), minimizing energy expenditure primarily through three main ways: adjusting the available ``supply'' or amount of resources $q_s$ (e.g. number/types of GPUs), adjusting resource allocation rules and schedulers $p$, and usage of control mechanisms $c$ (e.g. hardware settings). These mechanisms can be quite effective, cheap, and can easily produce intended results as they do not necessarily require coordination or know-how from users. While much work has focused on optimizing energy efficiency through these traditional mechanisms---affecting available compute resources, resource allocation and queuing/scheduling rules, or hardware/software and physical configurations \cite{frey2022benchmarking}\cite{carbonawarecompute}---new sources of efficiency will likely need to be claimed from $\varepsilon$ as we hit diminishing returns and, eventually, limits from traditional measures. As easy sources of efficiency are exhausted, these limits will require looking beyond more traditional levers (i.e., $p$, $q_s$, $c$) and towards less-traditional ones (i.e., $q_d$, $\varepsilon$).

\subsection{Energy, Power, \& Opportunity Costs}
When considering the energy expenditure or carbon footprint of HPCs/datacenters, what quantity should we focus on? As framed in Eq. \ref{energy_min}, the main objective $E(\cdot)$ can represent any number of quantities correlated with energy expenditure: kilowatt-hours, power usage effectiveness (PUE), pounds of CO$_2$ emitted, amount of water used in cooling, etc. Besides these quantities, $E(\cdot)$ can also account for aspects like the fiscal costs of the datacenter's energy bill or even the \textit{opportunity costs} of its choices, arising from the timing, the amounts, or the fuel composition of its energy demand and usage as well as how they affect the datacenter's environmental footprint. The economic costs of a choice accounts not only for its direct fiscal or monetary costs, but also its opportunity costs---the cost of the best alternatives foregone. In this subsection, we discuss these opportunity costs and strategies to reduce these costs by changing energy purchasing behaviors like the timing of energy purchases and other usage patterns. 

\begin{figure}[!h]
\centerline{\includegraphics[width=0.4\textwidth]{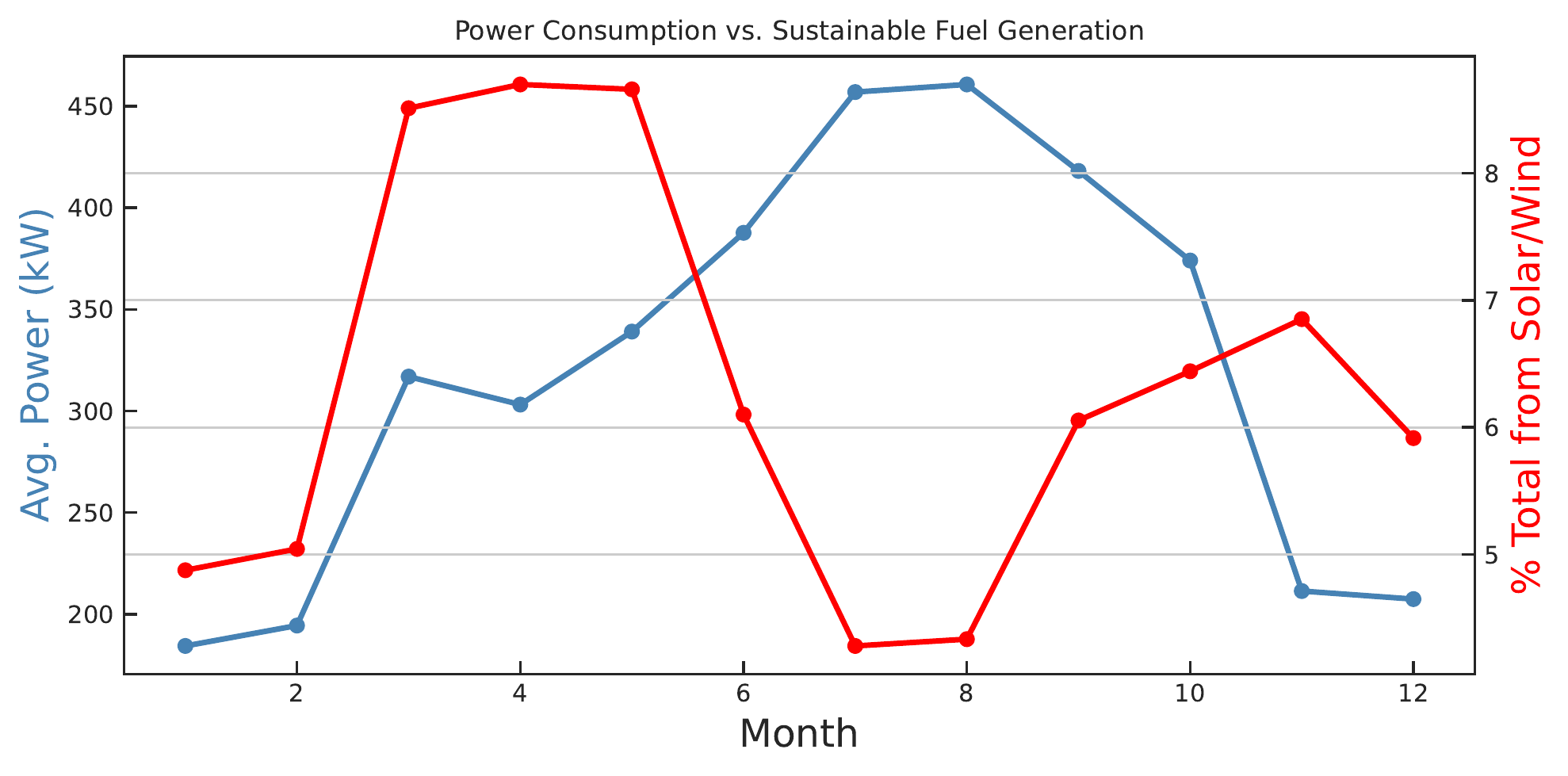}}
\caption{\small{\textbf{Power Consumption vs. Green Fuel Mix.} Average monthly power consumption of MIT's E1 hypercluster plotted against monthly average percentage of supplied total energy derived from solar and wind (2020-21). There are potential opportunities---high power consumption when green energy production is low and vice versa instead of the opposite.}}
\label{fig:fuelmix}
\end{figure}

\begin{figure}[!h]
\centerline{\includegraphics[width=0.4\textwidth]{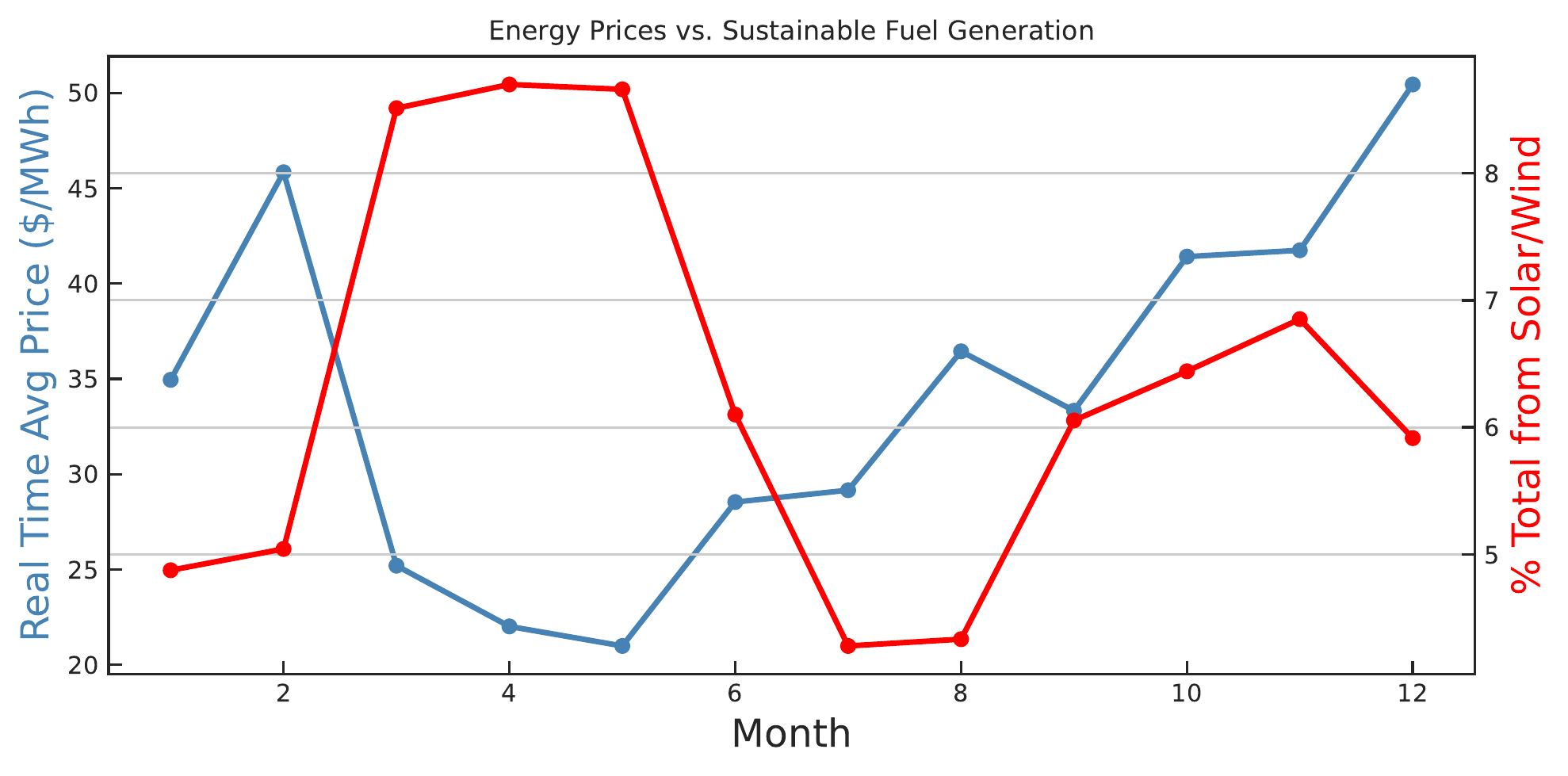}}
\caption{\small{\textbf{Energy Prices vs. Green Fuel Mix.} Average monthly energy prices plotted against monthly average percentage of supplied total energy derived from solar and wind (2020-21). Prices are monthly locational marginal prices (LMP) from south eastern/central MA. Note that energy prices tend to be lower when percentage of sustainable energy is higher.}}
\label{fig:prices}
\end{figure}

For instance, consider the usage patterns of the MIT SuperCloud system\cite{reuther2018interactive} within a given year. Naturally, the demand and usage of the system's overall resources will vary throughout a year, exhibiting regular patterns on different time scales within the year. Just as demand and load vary, power consumption will also vary---more users and jobs generally translate into more computation and increased cooling costs, increasing power draw from existing resources. Beyond the dollars-and-cents of the HPC's electricity bills, the make-up or composition of the energy supplied by the power company via the local grid can also influence the sustainability of a datacenter/HPC's operations albeit in a less direct way. The different sources from which power is generated (i.e. the fuel mix), supplied to, and consumed by the HPC carry an implicit environmental opportunity cost: the usage or purchase of power with a less sustainable fuel mix at a period in time forgoes usage of power generated with a greener fuel mix in that same time period. This, in turn, represents the foregone opportunity to offset some portion of existing energy expenditure while imposing an environment cost in the form of greater energy inefficiency as an externality. One way to then improve energy efficiency is to shift energy expenditure more towards power sourced from higher ratios of sustainable fuel mixes (i.e. generated with more sustainable sources like solar and wind).

Figure \ref{fig:fuelmix} suggests there may be an opportunity to change the datacenter/HPC's purchasing behavior for this strategy to be viable. Over the course of the year, we see that the total share of fuel/energy produced from solar and wind is inversely related to the average amount of power used per month. The MIT Supercloud energy consumption has been relatively high when the share of renewable energy is low around June to August---similarly, energy consumption/expenditure is lower when the share of renewable energy in the fuel mix of the power supplied is higher. One strategy to take advantage of this mis-match between power consumption and fuel mix, increase energy efficiency, and reduce the environmental opportunity cost is to purchase more power during times when sustainable energy takes up a larger share of the fuel mix (e.g. March to May) and either: (1) capitalize during that time period by encouraging more cluster utilization during those months or (2) store that energy to help offset energy consumption during times where the fuel mix is less sustainably sourced.

Figure \ref{fig:prices} suggests this strategy also carries financial benefits. During springtime, from February to May, when the sustainable energy share of fuel mix tends to be high ($> 8\%$), general energy prices tend to be extremely low (\$20-\$25 per megawatt-hour) and are some of the lowest prices of the year. However, it is important to note that renewable energies like solar and wind may not always see stable generation; moreover, there are additional fixed costs incurred from setting up the relevant infrastructure that may be required in order to pursue strategies like the ones described above. We explore and discuss the application of A.I. to help stabilize sustainable energy generation as well as infrastructure investments as they relate to efficiency in the sections below.

\subsection{Temperature-aware \& Weatherized Compute Optimization}

While changes in the regular, shorter-term behavior of datacenters/HPCs can be helpful, like those described above, longer-term structural changes and preparations are essential. As changes in climate produce increasingly extreme weather events and rising temperatures \cite{epaclimate}, traditional mechanisms alone may be insufficient to brace for what is to come. In light of these upcoming challenges, energy-aware cluster optimization must find ways to explicitly account for factors in $\varepsilon$ that, though difficult to anticipate, carry significant consequences to datacenter/HPC health and efficiency such as weather and climate. How would existing concepts and practices of cluster management and energy efficiency change with more extreme climate and more frequent weather events? What would weatherized compute optimization look like?

\begin{figure}[!htbp]
\centerline{\includegraphics[width=0.4\textwidth]{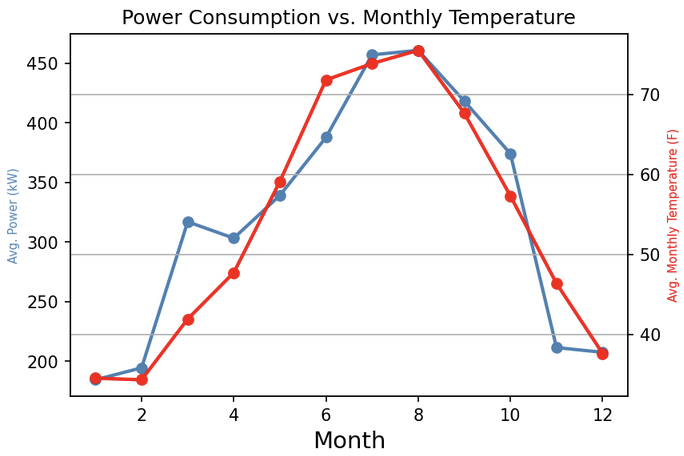}}
\caption{\small{\textbf{Power Consumption vs. Green Fuel Mix.} Average monthly power consumption of MIT Supercloud plotted against monthly average temperature (in Fahrenheit). Note the near one-to-one relationship between temperature and power consumption.}}
\label{fig:temp}
\end{figure}

In Fig. \ref{fig:temp}, we see the monthly average temperature and its trend along with those of power consumption for the MIT Supercloud system. Throughout the year, there is a monotonic, one-to-one relationship between average monthly power consumption and average monthly (local) temperature. As temperatures become warmer heading into the spring and summer months, it takes more power to cool the facilities and maintain a sufficiently low temperature for normal operations, resulting in increased power consumption. If average temperatures continue to climb even in the colder months as a consequence of climate change, cooling is likely to become more difficult and costly as previously efficient mechanisms for cooling facilities may suffer previously unseen stress.

As such, investments into infrastructure weatherization is critical. As changes in climate induce more extreme weather events and temperature ranges with increasing regularity, existing methods to realize energy efficiency may no longer be as effective under more frequent or extreme weather/climate conditions especially if mechanisms only function effectively within a small band of temperature/climate conditions. Since historical data points of extreme weather can be rare (for now), a useful exercise can be a regularly conducted stress-test akin to the Dodd-Frank stress tests \cite{fedres} enacted after the 2008 financial crisis; these stress tests are conducted annually and provide simulated stress scenarios that test the resiliency of financial institutions in both its traditional functions/operations as well as with less traditional risks (e.g. geopolitical, climate, infrastructure), helping identify areas in need of remediation. Similar stress scenarios and risk identification, conducted and evaluated regularly, for not just regular datacenter/HPC operations but also for climate and weather resiliency can help anticipate what energy efficiency (and inefficiency) looks like when considering future changes in weather and climate. For institutions with more than one HPC/datacenter, these exercises can provide opportunities to plan and coordinate across geo-scattered HPCs/datacenters to improve their collective resilience or develop re-routing backups in extreme weather conditions. Most importantly, these exercises can help anticipate and identify critical areas of infrastructure which require both a significant time and financial investment that may not come up otherwise.

\subsection{Incentives, Behavior, \& Mechanism Design}

Hardware and system-level mechanisms can carry much of the weight in producing energy savings under-the-hood and abstracting away difficulties without taking away from user experience. If these interventions run into diminishing returns, then discovering remaining gains in efficiency will require work not only from the ``supply'' side of computing but also on the ``demand'' side, $q_d$---the user. Compared to the macro-level approach dealing with cluster/datacenter-wide hardware and system-level interventions, this micro-level approach can provide additional flexibility but will require careful planning around mechanism design, user behavior, and user incentives. From this perspective, the optimization problem faced by the datacenter changes from Eq. \ref{energy_min} to
\begin{multline}
\label{energy_min_ind}
\min_{i} e_i( q_d(i), q_s, p, c, \varepsilon) \hspace{2mm} \text{s.t.} \hspace{2mm} a_i(q_d(i), q_s, p, c, \varepsilon) \geq \alpha_i
\hspace{2mm} \forall i \\ \text{where} \sum_i e_i = E, \hspace{1mm} \sum_i a_i = A
\end{multline}
for each individual or representative user (or workload) $i$. Whereas before the datacenter/HPC in Eq. \ref{energy_min} had control mainly through $q_s$, $p$, and $c$, now the main mechanism is through a specific user/profile/representative workload $i$. This ultimately translates into the datacenter attempting to induce changes in the quantity of resources demanded $q_d$, as reflected by $q_d(i)$. Instead of total across-the-board quantities like total energy and total activity/performance, $E(\cdot)$ and $A(\cdot)$, we now focus on individual (or representative) users, profiles, or representative workloads and their energy usage and activity profiles, as denoted by $e_i(\cdot)$, $a_i(\cdot)$, and $\alpha_i$. Naturally, by tailoring energy minimization efforts to representative user profiles and workloads, these mechanisms can reduce overall energy expenditure selectively in ways that systematic hardware interventions cannot. These micro-level approaches aim to induce behavioral changes in users through affecting incentives with the support of predictive analytics and instrumentation. %However, these mechanisms must monitor and balance the trade-offs between offering too much user control and offering too much system control to avoid resource mis-allocation.
    
One example is the design of queues for finer user and workload segmentation; these queues can improve job scheduling and execution using user-provided information (and other information) like the user's stated preferences on energy efficiency, job urgency/patience, expected time completion, type of workload, etc. Policies can then be tailored more specifically with only the resources necessary, allowing for more efficient design elements by reducing idle time, over-allocation, and over-utilization of resources. However, if queue selection and user intent conflict in situations where the user has an incentive towards a specific resource configuration different from the assigned one, this mechanism runs the risk of adverse selection---users mis-characterize their preferences and select themselves into queues where resources are fastest, most plentiful, or the most available, leaving select queues clogged and overtaxed and others largely, if not entirely, idle. 

In the example above, too many self-characterizing choices are made available for users to potentially mis-represent their preferences and extract private benefits while imposing a social cost on the whole system. One alternative to balance these two factors of too much choice and too little control is to maintain a two-part mechanism: a fixed component that guarantees a specified minimum amount of energy efficiency and a variable component that allows for user choice to further scale energy efficient behavior, but only in certain respects. For instance, it has been shown that optimal GPU power-caps provide an effective way to control energy consumption with minimal impact on training speed \cite{frey2022benchmarking} and user experience. With these optimal power caps as the fixed base component, the variable component can be offered as a choice: if an user accepts increasingly stringent power caps on his/her allocated GPUs (or other restrictions), the user can then, in exchange, choose to have more GPUs allocated to his/her tasks. These types of choice mechanisms require a cost-benefit analyses to balance individual net benefits/costs with system-level benefits/costs but can help induce energy-efficient changes in user behavior and computing demand. 

Designing mechanisms can be difficult but predictive models and analytical tools can help in understanding and evaluating both utilization patterns as well as opportunities to affect them in an energy-efficient way. Models that help forecast and relate energy prices, fuel mix, as well as energy expenditure to one another can provide significant support in the decision-making process for optimizing energy purchases and consumption. Similarly, models leveraging data on compute demand and usage (e.g. holidays, research deadlines) can help with scheduling, maintenance, etc. Though these mechanisms are not without their drawbacks, predictive analytics and instrumentation can help mitigate these shortfalls by anticipating and analyzing behavior via data and inference. 

\section{Climate-Aware Research Ecosystems}

A significant part of the A.I. research ecosystem is driven and structured by incentives to publish in notable, high-visibility conferences and journals. These venues serve as important forums for the A.I. community---researchers, practitioners, and the state of research as a whole---to disseminate new and important findings, promote brands, seek/hire talent, highlight significant contributions and problems, exchange information, foster innovation and collaborative relationships, and more. These contributions notwithstanding, the way the research ecosystem is currently structured can create incentives worth reconsidering when transitioning towards a more sustainable research environment. 

As both fundamental research and applications in A.I. to various fields continue to grow, high-visibility venues will likely receive more focus and submissions as researchers and practitioners strive to publish in the ``best'' possible venue. Many metrics of success in fundamental and applied research are also heavily influenced, if not defined, by publishing in these venues---preferring or requiring that researchers, practitioners, and even job candidates to have publications at notable venues---which continues to serve as a common incentive and evaluative criterion. With such a significant focus on publication in key conferences, how do these incentives drive the pattern of research activity and what environmental consequences do they carry, if any? Previous works have studied the carbon footprint generated by participants traveling to conferences \cite{carbonfootconf}\cite{globalwarmemission} but less attention has focused on the effect of the distribution of deadlines themselves.

\begin{figure}[!h]
\centerline{\includegraphics[width=0.4\textwidth]{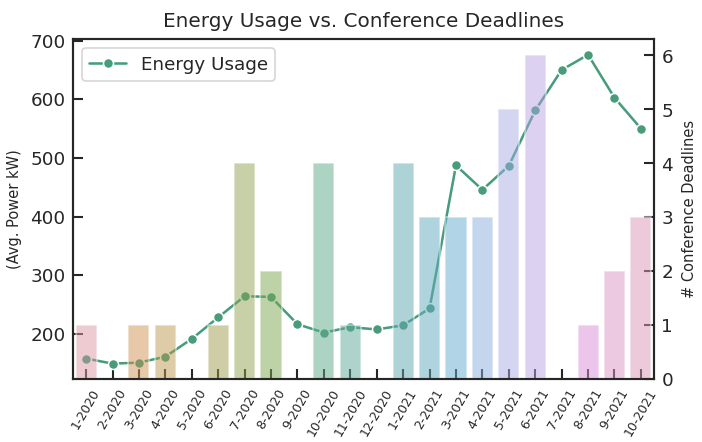}}
\caption{\small{\textbf{Energy Usage vs. Number of Conference Deadlines} Average monthly power consumption of MIT's E1 cluster plotted against number of monthly conference deadlines (Table \ref{tab:conference_list}) }}
\label{fig:confdead}
\end{figure}

Conferences deadlines are typically scattered throughout the year with each conference serving a specific domain or as a general purpose venue (e.g. see Table \ref{tab:conference_list}). Specific dates are publicized several months ahead to give enough time for preparation and planning. The distribution of these deadlines may induce certain patterns in aggregate research activity, compute demand, and therefore energy utilization, the last of which we use as a proxy for activity/demand. As an exploratory analysis, we compare the number of conference deadlines per month from January 2020 to end of year 2021 with trends in monthly energy usage in the MIT Supercloud system (Figure \ref{fig:confdead}). To help account for the confounding effects of seasonality, temperature, and other factors on energy utilization, we include data across two years (2021 \& 2022).

Given the way deadlines are structured, we might expect a lagging relationship where activity or compute demand, and hence energy utilization, might pick up in anticipation of upcoming deadlines---the larger the number or concentration of upcoming deadlines, the larger the increase in compute demand. As deadlines approach, users are accelerating their workloads, finishing or repeating experiments, and preparing for conference submission. In Figure \ref{fig:confdead}, we see some pick-up in energy usage leading up to the months with a high concentration of deadlines (i.e. July 2020)---such as the uptick starting around March/April 2020 and leading up to July 2020---but this may also be due to higher temperatures and cooling costs as noted earlier. However, there is a sharper pickup in energy usage starting around Jan/Feb 2021 in anticipation of a notable concentration of deadlines in the subsequent months. This sharp increase in energy usage is significantly higher than in the same period of the previous year despite no significant differences in average temperature or other known factors in those time periods between the two years---the only difference being the concentration/number of deadlines. Overall, we also see that many deadlines tend to concentrate in the spring/summer across both years when the combination of higher temperatures and increased compute demand can exacerbate existing energy trends, resulting in significantly higher energy usage that taxes the cluster. In the same period (i.e. the summer months), the fuel mix of the supplied power also has the lowest ratio of sustainable energy of the year, as seen earlier (Fig. \ref{fig:fuelmix}), which further contributes to an enlarged environmental footprint.

A natural question that may arise is: can we structure deadlines to spread out energy utilization and compute demand to benefit energy efficiency? If the same amount of compute is to be spent throughout an representative year of research activity regardless, then several options may help distribute that amount in a more sustainable fashion: (1) spread deadlines more uniformly throughout the year, (2) concentrate deadlines in the winter/spring months when preceding months are colder or see more sustainable fuel generation, or (3) abolish fixed deadlines in favor of rolling submissions. Some venues (e.g. Transactions on Machine Learning Research) have already shifted to rolling submissions albeit for different reasons. 

We note that our preliminary analysis is intentionally limited in scope as we focus exclusively on the MIT Supercloud system. Additionally, it neither accounts for other confounding factors explicitly nor does it show a definitive connection between conference timings and usage/energy intensity. Rather, it is meant to bring attention to how structural incentives in the current A.I. research ecosystem and community may not align optimally with desirable aspects of sustainability---with one example being conference deadlines. More work and data are required to tease out the full picture of the degree to which aggregate research activity and its energy footprint are affected by conference timings. We hope that future work will undertake a finer analysis, accounting for details such as workload type, type of research activity represented, breakdown of activity and energy use by domain (e.g. NLP), etc. beyond just data from this cluster. This requires more data, better data, data access, as well as willingness to share these data, which may not currently exist in sufficient amounts, a matter we discuss further below.

\begin{table}[!htbp]
\caption{\textbf{List of notable conferences.} The following conferences are considered for analysis (not exhaustive).}
\begin{center}
\begin{tabular}{cll}
\toprule
\textbf{Area/Discipline}&\textbf{Conferences}& \\
\midrule
NLP/Speech & EACL, InterSpeech, EMNLP, AKBC, ICASSP & \\
& ISMIR, AACL-IJCNLP, COLING, CoNNL,& \\
&  WMT, EACL & \\
\midrule
Computer Vision & ICME, ICIP, SIGGRAPH, MIDL, ICCV, & \\
& FG, ICMI, BMVC, WACV & \\
\midrule
Robotics  & IROS, RRS, CoRL, ICRA & \\
\midrule
General ML  & COLT, ICCC, ICPR, AAMAS, AISTATS, CHIL & \\
&  EMCL-PKDD, NeurIPS, ACML, AAAI, ICLR& \\
\midrule
Data Mining  & SDM, KDD, SIGIR, RecSys, CIKM, ICDM & \\
& WSDM, WWW & \\
\bottomrule
\end{tabular}
\label{tab:conference_list}
\end{center}
\end{table}

\section{Climate-Aware Research Priorities}
A discussion on the sustainability of the current A.I. research ecosystem and its incentives would be incomplete without discussing the thematic lines of work, both old and new, such an ecosystem should prioritize in order to improve its sustainability and keep its environmental footprint small.

\subsection{Novelty, Redundancies, \& Efficiency}
Given the complexity and variety of research and applications in A.I., there are likely significant redundancies in A.I. workflows. Many experiments usually begin with training known and proven models up to some pre-specified level of performance, depending on the research direction, before building atop these results. Doing so may require some hyper-parameter search, if not full-blown optimization, resulting in multiple training runs and inevitably redundant runs, wasted compute, and additional energy costs. Some redundancies can play a helpful role by training students and researchers when they start working on A.I. research where experience obtained from reproducing results can help shape best practices down the road. However, problems with reproducability of research only compound these redundancies as (multiple) attempts at replication also waste resources and energy when researchers and practitioners attempt to build off existing work or put previous work into practice. These difficulties in replicating published results are wide-spread and well-documented\cite{repro}, resulting from inconsistent reporting of sensitivity to hyper-parameters and training settings (or complete lack thereof), poor communication, missed opportunities from reviewers, mis-representation, or some combination of the above.

In the ever-changing landscape of new research and model frameworks, problems with redundancy and reproducibility can carry additional implications for energy efficiency. If incentives to develop better performing models overshadow those for reproducibility and transparency, research efforts devoted to producing newer, better models will outpace efforts for clearer benchmarking and reporting, leaving transparency and resource efficiency efforts forever playing catch-up. For instance, when GPT-3 debuted, despite its impressive performance on generative language tasks, its training (not including experimentation during its development) was prohibitively costly and estimated at around \$5 million using a specially designed supercomputer by Microsoft\cite{microsoft}, making it very difficult for researchers to train and test on their own---only after its introduction, extensive usage, and popularization did work focus addressing its efficiency and other issues (e.g. safety, A.I. alignment, etc.). Over-parameterization and big data may offer easy performance improvements, but an emphasis on jointly co-optimizing efficiency and performance in research may help avoid this efficiency-in-hindsight approach and front-loading significant energy costs in model development. Some progress has been made in addressing these problems as Google, Meta, and other large players have highlighted best practices and standards that have helped to significantly reduce their own carbon footprints \cite{google2022carbon}\cite{metasusAI} for state-of-the-art NLP models, such as efficient model selections and hardware/system choices. Despite this, however, the fundamental problem of information reporting and data availability still remains. To remedy this, there needs to be an active, systematic, and consistent approach towards collecting and reporting data/information (on energy usage, training settings, etc.) that incentivizes voluntary contribution and surveys a sufficiently broad swath of sources to be representative of the diversity of workloads in research and practice.

% New development frameworks may offer more functionality and accessibility by abstracting technical aspects away from user experience, but how well they scale or train in-line with existing guidelines and rules-of-thumb regarding efficiency under different compute settings is less clear, raising questions on their energy/computational efficiency. 

\subsection{Measurement, Reporting, \& Transparency}
Various works have produced estimates in attempts to quantify the carbon or energy footprint of deep learning model training with estimates ranging from as high as 5x the average lifetime emissions of a car \cite{strubell2019energy} to as low as $10^{-5}$ times that amount \cite{google2022carbon} for state-of-the-art transformers. These estimates are inherently variable and difficult---not only due to differences in aspects like hardware (e.g. GPU vs. TPU)---in both the approach taken to quantify these costs and their resulting accuracy. These difficulties in accurate estimation highlight the importance of regularly detailing energy usage and other information in research alongside typical items like performance results and ablation tests. Moreover, while many estimates have focused on training costs, even less clear are the costs arising through a model's entire life-cycle, which are particularly important in industry and applied settings. Even so, there exist even less data on the costs of inference. 

The discrepancies in, and even availability of, these estimates can be due to several reasons. The first is resource asymmetry---not only do different companies, groups, and individuals have different amounts of computational resources, they also have different computational setups so certain metrics and calculations may naturally vary depending on the underlying technological stack. This differentiation similarly applies in academic disciplines where a base model (e.g. graph neural networks) may branch out into highly specialized, differentiated variants depending on the field or task (e.g. social networks vs. molecular predictions), resulting in significantly different training procedures, learning dynamics, energy footprints, and more. Different needs, resources, and constraints largely determine variations across research and development workflows; as such, when a company or institution reports realized gains in efficiency or savings, \textit{these gains may only be realizable on their systems, with their resources/hardware/configuration, or limited to a specific class of models that are reported by, or essential to, said organization}. Though a seemingly simple solution would be to move over to services provided by organizations with the hardware and technical capabilities to realize such efficiencies, there are ethical concerns and market concentration issues that require addressing. Even with similar tasks across companies and industries, different domains are also characterized by other considerations and constraints such as the lack of technical expertise, specific resource and regulatory constraints, and other requirements like model privacy or interpretability that may outweigh model performance and efficiency. At its worst, resource asymmetries can hamper reproducibility and verification efforts: if state-of-the-art models developed by large, well-equipped research groups are too costly and resource-intensive to train for others, how can their results and estimates be reproduced or verified? 

Along with the resource asymmetry, information asymmetry can discourage and dis-incentivize researchers and practitioners from reporting necessary or relevant information. Some examples of these asymmetries, besides ones mentioned earlier like inconsistent reporting of training
settings as well as poor communication and presentation of research results, can arise in part from incentives to preserve competitive advantages and other sensitive information. Incentives to protect and preserve a competitive edge from peers and competitors can discourage full, transparent reporting of information especially if these models and research tie into a company's products and services. Even when reporting, these incentives may limit the amount of information made available to the wider research community, leading to confusion around estimates and methodologies. Incentives to keep information, and its benefits, private for competitive advantage can lead to continued information asymmetries in a self-reinforcing cycle. Voluntary reporting may then be dominated by larger, better-equipped groups with the resources and technical ability to optimize their operations which, though well-intentioned, will likely not reflect the true extent of the overall, or even the average, environmental footprint of A.I. and its applications. Moreover, despite the focus on the footprint and costs of training, data and estimates on inference are even scarcer despite its significance---the few estimates, where available, put inference at 90\% of production ML infrastructure costs\cite{awscompute} and 80\%-90\% of energy costs\cite{nvidiainfercosts}. While training enjoys scaling benefits that saturate GPUs, the different performance requirements of inference can result in poor GPU utilization since inference queries are unable to realize the parallelism that offline mini-batch training enjoys \cite{gpuinference}. Low resource-efficiency and utilization is quite common: AWS reports p3 GPU instances at only 10\%-30\% utilization\cite{awscompute} and even Google's TPUs exhibit a utilization of 28\% on average \cite{googlecomputeutil}.

The issues outlined above all point to a common set of problems that require (1) a better, more representative idea of the kind of A.I. models, and the underlying resources, used across disciplines, domains, and communities, (2) a common set of meaningful metrics, and (3) incentives through both existing avenues (e.g. conferences, papers) and new ones such as forums, competitions, leaderboards, or open challenges to encourage reporting of energy/utilization data and development of more energy-efficient models rather than just better performing ones. To accurately quantify the environmental footprint, it is essential to capture costs with metrics that realistically reflect and represent the workloads undertaken in A.I. research and practice---as well as the burdens and energy footprint associated with state-of-the-art models on more representative computational setups rather than in the most efficient, advanced settings. To incentivize consistent reporting and sharing of data, the research community needs forums that prioritize energy-efficient models and methodologies. For instance, a Green A.I. challenge (in development) that aims to cast the problem explicitly by challenging participants to maximize performance given explicit training and energy budgets. Lastly, facilities should also provide the central infrastructure, user interfaces, and analytical tools/instrumentation/logging to further encourage easy reporting and sharing of data, especially since not all users are equipped with the expertise to manually report relevant data and information.

\subsection{A.I. for  Energy Savings, Generation, \& Discovery}
Despite its potential environmental footprint, some of the most impressive applications of A.I. algorithms have included ones that help generate energy savings themselves. One example has been Google and DeepMind's use of neural networks to monitor and optimize their datacenters, reducing the amount of energy spent for cooling by 40\% and PUE by 15\% in live tests\cite{deepmindcooling}. Similar examples abound, but beyond energy savings, continued and improved sustainability will also require work from the other side of the equation: energy generation.

The study and application of A.I. to energy discovery and generation should be strongly incentivized given its immediate benefits. Current examples include the application of algorithms to stabilize and boost sustainable energy generation: wind farms provide inexpensive, carbon-free energy but can be unpredictable, making planning and energy delivery/storage difficult. In response, DeepMind has developed neural networks trained on weather forecasts and historical turbine data to forecast energy output 36 hours ahead, making early recommendations on optimal hourly delivery commitments to the grid possible \cite{deepmindenergygen}. Beyond existing energy sources, A.I. research can help push forward new sustainable energy sources. Recent work has shown how deep reinforcement learning can help control nuclear fusion \cite{RLnuclear} by learning to control and change the shape of plasma via manipulation of its magnetic field. Scientific collaborations, especially as they relate to development of new energy sources or improvements in existing energy generation, should receive equal priority and recognition as state-of-the-art performance improvements in areas like vision and NLP. To do so, partnerships with scientific and energy researchers should be encouraged and made more accessible to A.I. researchers and practitioners. Similarly, benchmark energy datasets should be constructed and made easily accessible just like standard data benchmarks in NLP and vision---moreover, these energy datasets should receive continuous updates and testing due to the inherently variable behavior of wind, weather, etc.  

\section{Conclusion}

There are many dimensions of this multi-faceted problem that are not addressed in this paper due to space limitations but are important for consideration nonetheless such as the equity and accessibility aspects of energy-efficient computing. Though daunting, we hope our discussions of these problems and their potential solutions will provide a framework that spurs further discussion, and most importantly action, on these various issues.
% One such issue is the aspect of equity and accessibility in A.I. research. Given the relatively high barrier to entry in the form of access to significant computational resources, some areas of research (e.g. protein folding, reinforcement learning, neural architecture search) can be near inaccessible for some---ultimately limiting the novelty and diversity of ideas, as well as representation, in research. This in itself can foster a vicious cycle where the ability to make state-of-the-art advances are effectively monopolized by large, well-equipped research groups and companies.
% To truly understand the scale and scope of the environmental footprint of A.I. research, we need accurate, consistent reporting of information in the form of a commonly defined set of metrics and ways to better incentivize consistent reporting, transparent research and communications, and information disclosure.

\section*{Acknowledgment}
The authors acknowledge the MIT SuperCloud \cite{reuther2018interactive} and Lincoln Laboratory Supercomputing Center for providing HPC and consultation resources that have contributed to the research results reported within this paper.
The authors acknowledge the MIT SuperCloud team: William Arcand,  William Bergeron, Chansup Byun,  Michael Houle, Jeremy Kepner, Anna Klein, Peter Michaleas, Lauren Milechin, Julie Mullen, Albert Reuther, Antonio Rosa, and Charles Yee. The authors also wish to acknowledge the following individuals for their contributions and support: Bob Bond, Allan Vanterpool, Tucker Hamilton, Jeff Gottschalk, Tim Kraska, Mike Kanaan, Charles Leiserson, Dave Martinez, John Radovan, Steve Rejto, Daniela Rus, Marc Zissman.

\balance
\bibliographystyle{IEEEtran}
\bibliography{IEEEabrv,bib}

\end{document}